\documentclass[letterpaper, 10 pt, conference]{ieeeconf}  % Comment this line out if you need a4paper

\IEEEoverridecommandlockouts                              % This command is only needed if 
\usepackage{graphics} % for pdf, bitmapped graphics files
\usepackage{graphicx}
\usepackage{amsmath} % assumes amsmath package installed
\usepackage{bm}
\usepackage[noadjust]{cite}
\usepackage{amssymb}  % assumes amsmath package installed
\usepackage{algorithm}
\usepackage[noend]{algpseudocode}
\usepackage{svg}
\usepackage{soul}
\usepackage{multirow}
\usepackage{makecell}
\usepackage{hyperref}

\allowdisplaybreaks[4]
\def\BibTeX{{\rm B\kern-.05em{\sc i\kern-.025em b}\kern-.08em
    T\kern-.1667em\lower.7ex\hbox{E}\kern-.125emX}}
\usepackage{todonotes}
\title{\LARGE \bf
Diffusing Trajectory Optimization Problems for Recovery During Multi-Finger Manipulation
}

\author{
  Abhinav Kumar$^{1}$, Fan Yang$^{1}$, Sergio Aguilera Marinovic$^{2}$, Soshi Iba$^{2}$,  Rana Soltani Zarrin$^{2}$, Dmitry Berenson$^{1}$\\
  \thanks{
  $^{1}$Robotics Department, 
        University of Michigan, Ann Arbor, MI, USA
        {\tt\small abhin@umich.edu},  $^{2}$ Honda Research Institute USA. This work was sponsored by Honda Research Institute USA.}
  }

\begin{document}
\everypar{\looseness=-1}
\newcommand{\cmode}{\mathbf{c}}
\newcommand{\sinit}{\mathbf{s}_0}
\newcommand{\state}{\mathbf{s}}
\newcommand{\control}{\mathbf{u}}
\newcommand{\obj}{\mathbf{o}}
\newcommand{\traj}{\bm \tau}

\newcommand{\abhinav}{\textcolor{black}}
\newcommand{\namenospace}{D-TOUR}
\newcommand{\name}{\namenospace~}
\def\thickhline{\noalign{\hrule height.8pt}}
\maketitle
\thispagestyle{empty}
\pagestyle{empty}
%===============================================================================

\begin{abstract}
    Multi-fingered hands are emerging as powerful platforms for performing fine manipulation tasks, including tool use.
    However, environmental perturbations or execution errors can impede task performance, motivating the use of recovery behaviors that enable normal task execution to resume.
    In this work, we take advantage of recent advances in diffusion models to construct a framework that autonomously identifies when recovery is necessary and optimizes contact-rich trajectories to recover.
    We use a diffusion model trained on the task to estimate when states are not conducive to task execution, framed as an out-of-distribution detection problem.
    We then use diffusion sampling to project these states in-distribution and use trajectory optimization to plan contact-rich recovery trajectories.
    We also propose a novel diffusion-based approach that distills this process to efficiently diffuse the full parameterization, including constraints, goal state, and initialization, of the recovery trajectory optimization problem, saving time during online execution.
    We compare our method to a reinforcement learning baseline and other methods that do not explicitly plan contact interactions, including on a hardware screwdriver-turning task where we show that recovering using our method improves task performance by 96\% and that ours is the only method evaluated that can attempt recovery without causing catastrophic task failure.
    \abhinav{Videos can be found at \href{https://dtourrecovery.github.io/}{https://dtourrecovery.github.io/}.}
\end{abstract}

% Two or three meaningful keywords should be added here
% \keywords{In-Hand Manipulation, Contact-Rich Manipulation, Learning for Planning} 

%===============================================================================

\vspace{-.2cm}
\section{Introduction}
\vspace{-.2cm}
\label{sec:introduction}
As progress is made on techniques that enable fine multi-finger manipulation \cite{kumar2024diffusion, fanrolling, pang2023global, jiang2024contact, chen2023visual, wang2024penspin, kurtz2023inverse}, it is important to consider scenarios in which these methods might perform poorly.
These could occur due to execution error or external perturbations.
For example, for precise tasks such as turning a screwdriver, perturbations to the object could push the system state out of the domain in which the task policy was trained.
By detecting these scenarios, we can execute recovery behaviors after which the system state is conducive to normal task execution.
Prior work addresses learning recovery behavior \cite{zhou2023adaptiveonlinereplanningdiffusion, thananjeyan2021recovery, wang2019learning, luo2021endowing, reichlin2022back, 813040, vats2024plan}.
In contrast to prior work, we focus on learning recovery behaviors that are less likely to lead to catastrophic failure by imposing strict constraints using trajectory optimization, where we generate the trajectory optimization problem and initializations online using a learned model.

Our method, Diffused Trajectory Optimization for Use in Recovery (\namenospace), uses a diffusion model \cite{ddpm, song2020score} to generate trajectory optimization problems solved to execute recovery behavior.
The model samples the full parameterization of a trajectory optimization problem, including constraints, objectives, and initializations.
% We build off prior work on estimating likelihoods of trajectories under diffusion models \cite{zhou2023adaptiveonlinereplanningdiffusion}, using it to evaluate how likely a diffusion model is to sample reasonable trajectories given an initial state.
% We use this likelihood as an OOD score of the state, using it to detect when recovery is needed.
% A key insight is to use the innate likelihood-maximizing properties of diffusion sampling to project OOD states in-distribution, creating a goal state for recovery.
% We propose a method to do this that directly uses a trajectory diffusion model, without requiring training a separate model of the state distribution.
% We then use a contact-rich trajectory optimization approach to generate recovery trajectories.
% We then distill this recovery behavior into a novel diffusion model approach that diffuses the full parameterization of a recovery trajectory optimization problem, including constraint, goals, and initializations to improve planning speed online.
Starting with a separate diffusion model trained on trajectories from normal task execution, we execute the task in simulation and apply perturbations.
We use the task diffusion model to detect states from which it does not sample useful trajectories, taking advantage of recent work in estimating likelihoods of diffused trajectories \cite{zhou2023adaptiveonlinereplanningdiffusion} to structure this as an out-of-distribution (OOD) detection problem.
% We construct a dataset of recovery trajectories by executing a task, perturbing the system during task execution.
% Given a diffusion model trained on successful task executions, we use existing techniques for estimating trajectory likelihoods \cite{zhou2023adaptiveonlinereplanningdiffusion} to detect states from which the diffusion model cannot sample useful trajectories.
We initiate recovery when we encounter OOD states.
To inform generation of a training dataset of optimization problems, we propose a novel method to project OOD states in-distribution (ID) of the task diffusion model.
This ID state serves as the target for recovery.
We rely on a key insight, which is to use the natural likelihood-maximizing properties of diffusion sampling to perform the projection.
To further structure the trajectory optimization, we define contact modes which encode contact behavior for the fingers of a robot hand.
These contact modes define constraints and objective functions, where specific parameters of these functions are calculated from the projected ID state.
The recovery diffusion is then trained to jointly sample contact modes and recovery trajectories, where the joint sampling allows us to reconstruct a recovery trajectory optimization problem online.
Additionally, we use the sampled trajectories as initializations to the non-convex optimization.
In this work, we use CSVTO \cite{power2024constrained} to solve the optimization, though other solvers could also be used.

\begin{figure}[t]   
% \centerline{\includegraphics[width=\textwidth]{figures/block_diagram.pdf}}
% \vspace{-.55cm}
\centerline{\includegraphics[width=\linewidth]{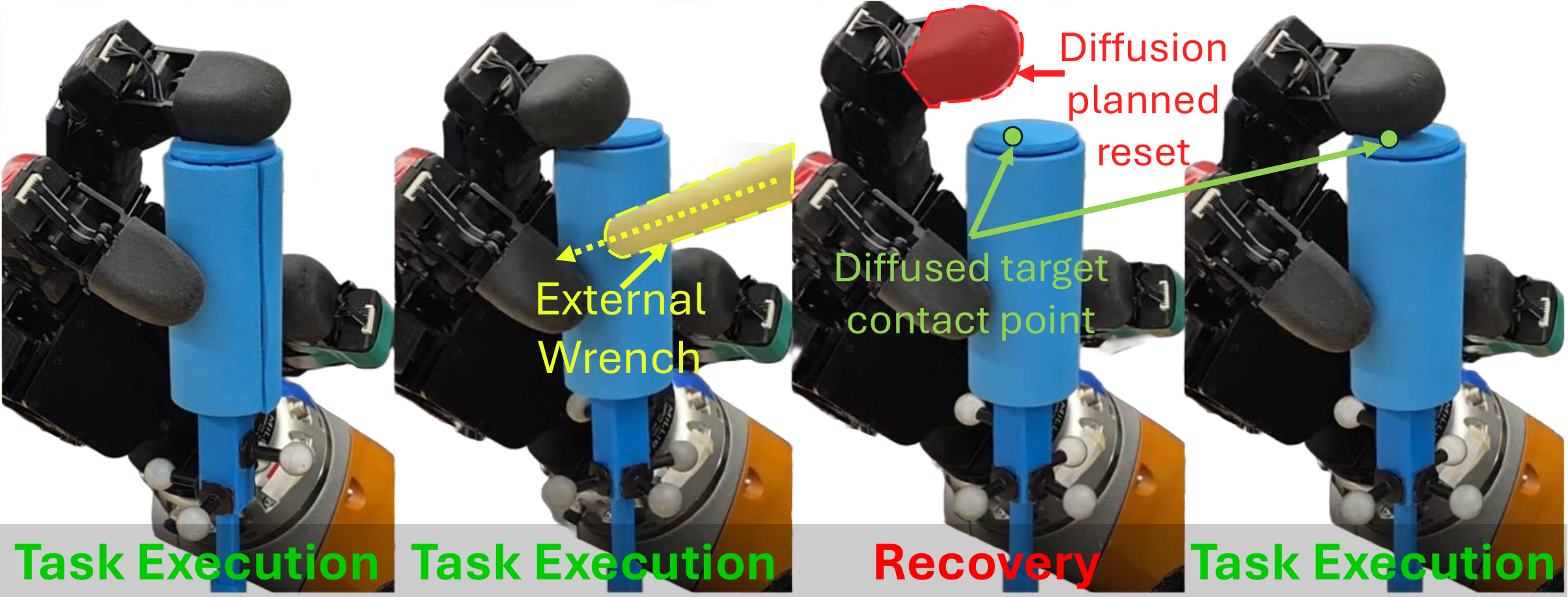}}
\vspace{-.3cm}
\caption{During a screwdriver turning task, we apply external wrench perturbations.
Our method detects that recovery is needed and diffuses a trajectory optimization problem that encodes the set of fingers to reset and corresponding target contact points. The index finger is reset in this example.}
% \vspace{-.2cm}
\label{fig:TODR_title_figure}
\end{figure}

Our contributions are (1) A framework that uses diffusion models to generate recovery behavior and switch between task execution and recovery execution as needed; (2) A method, based on diffusion sampling, to project OOD states in-distribution and inform recovery trajectories; (3) A method to diffuse trajectory optimization problems that are solved to realize in-distribution projection of states.

We evaluate our method on a valve turning task and a screwdriver turning task in simulation and on hardware. 
Our results show that our method allows us to detect when recovery is necessary and execute recovery behavior in a way that improves task performance.
We compare against a safe reinforcement learning baseline and methods that do not diffuse the full optimization problem, or do not reason about specific contact interactions. 
We show these methods are more likely to lead to catastrophic task failure and less likely to recover in a way that aids task execution.
We also compare against multiple methods that do not explicitly reason about the difference between task execution and recovery or explicitly reason about different contact modes, showing the benefit of explicit reasoning.

%===============================================================================
\vspace{-.2cm}
\section{Related Work}
\vspace{-.1cm}
\label{sec:related_work}

\textbf{Recovery:} Prior work has examined how to detect and recover from problematic states \cite{zhou2023adaptiveonlinereplanningdiffusion, thananjeyan2021recovery, wang2019learning, luo2021endowing, reichlin2022back, 813040, vats2024plan}.
Some works using reinforcement learning (RL) \cite{thananjeyan2021recovery, wang2019learning, luo2021endowing} learn classifiers that detect when recovery is required and recovery policies to compute the recovery behavior.
The tasks we consider have contact constraints that induce low-dimensional constraint-satisfying trajectory manifolds and these methods can struggle to learn policies that avoid catastrophic failure, i.e. dropping a tool.
We address this issue using a combination of diffusion, which has been shown to learn useful models of high-dimensional distributions, and trajectory optimization to compute recovery behaviors which allows us to better satisfy constraints that are important for enabling fine manipulation.
We compare against \cite{thananjeyan2021recovery} in our experiments to show the importance of explicitly considering constraints in recovery.%showing our diffusion-based framework outperforms their proposed RL method.

Other works use behavior cloning methods for task policies and use density estimation to decide when recovery is needed, using gradients of density estimates to compute recovery actions \cite{reichlin2022back}.
We also use a variant of density estimation through diffusion model likelihoods and train a recovery policy that is informed by these density estimates.
However, our recovery policy is of a more general form than the one in \cite{reichlin2022back} which defines actions only as changes in state and directly executes the gradient of the state density estimation network.
In multi-finger manipulation tasks, it can be difficult to follow a direct state gradient to recover, requiring more complex recovery policies.

\textbf{Implicit Methods:} In contrast to recovery methods, other works approach the full manipulation problem with a single policy that implicitly reasons about contact mode changes.
These include reinforcement learning methods such as \cite{qi2023hand, chen2023visual, wang2024penspin}, and contact-implicit trajectory optimization methods such as \cite{pang2023global, jiang2024contact, kurtz2023inverse}.
While these methods are attractive due to their ability to implicitly choose when to switch contact and react to disturbances, they provide less control and interpretability over when these behaviors occur, which is problematic for fine manipulation tasks.
We compare against \cite{qi2023hand} and \cite{kurtz2023inverse} to illustrate the benefits of our explicit reasoning over recovery and contact modes.
% \textbf{Diffusion for Manipulation}

\textbf{Diffusion and Planning:} Diffusion models have recently emerged as powerful planners.
Prior work that combines trajectory optimization or planning with diffusion usually focuses either on using diffusion models to initialize non-convex optimization problems \cite{kumar2024diffusion, ortizharo2021structured, huangdiffusionseeder} or using diffusion models to solve trajectory optimization or motion planning problems with \textit{a priori} known constraints and objectives \cite{trajdiffuser, kondo2024cgd, carvalho2023mpd, pan2024model, liang2023adaptdiffuser, mishra2023generative, mishragenerative, huang2023diffusion, yan2025m}.

Additionally, prior work has explored using planning to generate trajectory data for diffusion models \cite{li2024planning, yang2025physics} as we do, but we diffuse trajectory optimization problems along with trajectories to ensure constraint satisfaction of trajectories. 
While prior work has learned cost functions \cite{urain2023se} and constraint representations for collision-checking \cite{9561516}, we focus on diffusing the entirety of the optimization problem.
% In addition, we do not use implicitly learned constraints or objectives to improve interpretability and reduce approximation error. 
% In add?ition, we use explicit 
\cite{weng2024dexdiffuser} diffuses grasps for multi-finger hands.
While this could compute ID states in our framework, it trains an additional diffusion model and classifier to perform the sampling unlike our method which only uses the task diffusion model.

% This allows us to simultaneously plan contact behavior at a higher-level, i.e. make decisions on which fingers should be used in a contact-rich manipulation task, while also solving a lower-level optimization problem to compute 

\begin{figure*}[t]   
\centerline{\includegraphics[width=\textwidth]{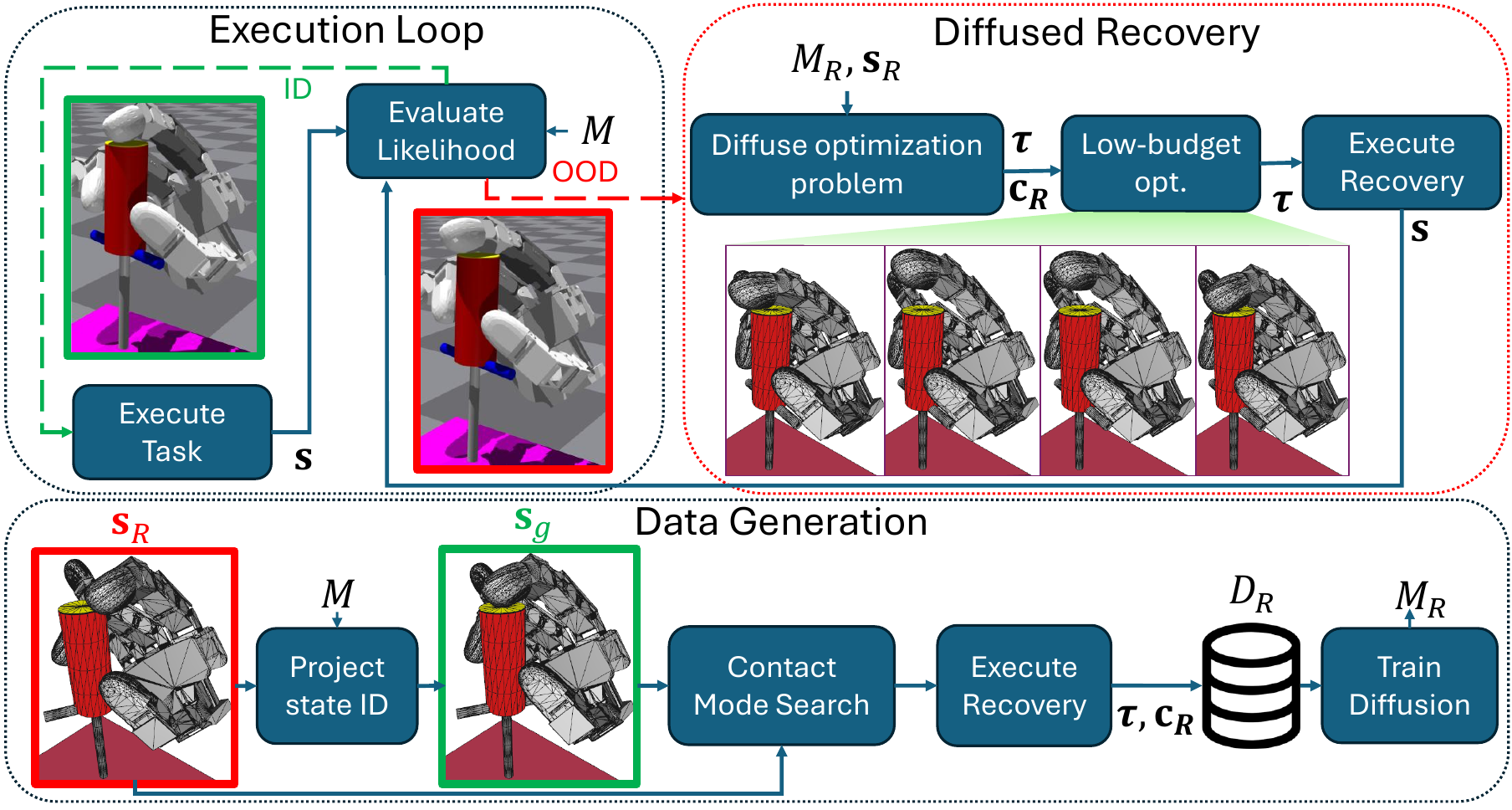}}
% \centerline{\includegraphics[scale=.5]{figures/TODR_block_diagram.pdf}}
\vspace{-.4cm}
\caption{While executing a task, our method uses a task diffusion model $M$ to detect when the current state is out-of-distribution (OOD).
We initiate recovery behavior in OOD states as indicated by the dashed red line.
We train a diffusion model $M_R$ to jointly diffuse trajectories $\traj$ and contact modes $\cmode_R$ which together parameterize and initialize a trajectory optimization problem.
% We additionally use $\traj$ to initialize the trajectory optimization, allowing us to use a low budget for optimization.
We resume execution after recovery if the state is ID, indicated by the dashed green line. Otherwise, we retry recovery.
To generate training data, we first project OOD states $\state_R$ to ID $\state_g$ using $M$.
We then search over a set of contact modes, choosing the one that leads to the highest likelihood state.
We add the planned trajectory and contact mode to the dataset $D_R$ and train $M_R$ on this dataset.}
% \centerline{\includegraphics[scale=.95]{figures/projection_screwdriver_only.pdf}}
% \caption{Our method using diffusion sampling to project an out-of-distribution initial state $\sinit$ in-distribution.
% By iterating over multiple contact modes $\cmode_i$, where $\cmode_i$ defines a trajectory optimization, we compute a candidate recovery trajectory. We select the recovery trajectory that terminates with the highest likelihood state and add it to our dataset $D_R$ along with its contact mode $\cmode_R$.
% We train a diffusion model $M_R$ to jointly diffuse $\traj, \cmode_R$, effectively diffusing a trajectory optimization problem.
% Online, we sample a trajectory optimization problem from $M_R$ to recover, additionally using $M_R$ to sample initializations to the optimization.}
\vspace{-.8cm}
\label{fig:block_diagram}
\end{figure*}	
%===============================================================================
\vspace{-.3cm}
\section{Problem Statement}
\vspace{-.2cm}
\label{sec:problem_statement}
Within the broader problem of manipulation with multi-finger hands, we focus on detecting and recovering from states that, if recovery were not performed, would inhibit task progress.
We define the state $\mathbf{s}_t$ at time $t$ as $\{\mathbf{q}_t, \mathbf{o}_t\}$, where $\mathbf{q}_t$ is the robot configuration and $\mathbf{o}_t$ is the object configuration, which takes different forms depending on the object.
We assume the system is quasi-static, meaning that all velocities are 0 before each action is taken.
We additionally assume access to a simulator where we can model the system and execute the task.
To enable this, we assume that geometries of the object and robot are known.
This would be the case in a factory setting, where robots would be manipulating a known set of objects.

As part of executing recovery behaviors, we choose how fingers make and break contact with the object.
For example, we may want to reset the position of a specific finger that has slipped off a screwdriver while attempting to turn it.
To help with this, we reason over contact modes that encode the contact behavior of each finger. 
% \abhinav{considering a single contact point per finger}.
We define contact mode $\mathbf{c}:=\{0, 1\}^{n_f}$, where $n_f$ is the number of fingers.
% $\mathbf{c}$ is a binary vector specifying which fingers should remain in contact when executing a trajectory.
A value of 1 means the finger should remain in contact throughout the trajectory, whereas a value of 0 means the finger should break contact with the object before making contact again.
We refer to this behavior as ``resetting" the finger.

Contact modes define inequality constraints $g$, equality constraints $h$, and objectives $J$ for a trajectory optimization problem that produces recovery behavior.
The trajectory optimization problem, shown in \eqref{eq:ps_trajopt}, optimizes trajectories $\traj:= \{\mathbf{s}_{0:H}, \mathbf{u}_{0:H-1}\}$ of length $H$, where $\mathbf{u}_t$ is a control input $\{\Delta \mathbf{q}_i, \mathbf{f}_i\}^{n_f}_{i=1}$. 
\abhinav{$\Delta \mathbf{q}_i$ is the commanded change in $\mathbf{q}$ and }$\mathbf{f}_i$ is the force applied by the $i$th finger.

\vspace{-.5cm}
\begin{align}
\begin{split}
    \mathbf{s}^*_{0:H}, \mathbf{u}^*_{0:H-1} &= \arg \min J(\mathbf{s}_{0:H}, \mathbf{u}_{0:H-1}, \mathbf{c}) \\
    &\text{s.t.} \quad   
    h(\mathbf{s}_{0:H}, \mathbf{u}_{0:H-1}, \mathbf{c}) = 0 \\
    &g(\mathbf{s}_{0:H}, \mathbf{u}_{0:H-1}, \mathbf{c}) \leq 0,
\end{split}
\label{eq:ps_trajopt}
\end{align}
\vspace{-.3cm}

We assume access to a task diffusion model $M$ trained on task executions, as done in \cite{kumar2024diffusion, zhou2023adaptiveonlinereplanningdiffusion, mishra2023generative, mishragenerative}.
$M$ can either be directly used during task execution or can model the distribution of trajectories output by the task policy.
% We also assume access to the dataset $D$ that $M$ was trained on and that $M$ can provide estimates on the likelihoods of trajectories.
We assume $M$ can be conditioned on $\mathbf{s}_0$ so that all sampled trajectories have $\sinit$ as the initial state.
We use this diffusion model to calculate the likelihood of states encountered during task execution and inform recovery trajectory generation.

We evaluate our method on its ability to recover from detected errors as well as overall task performance to evaluate if recovery is beneficial for task execution.

%===============================================================================

\vspace{-.3cm}
\section{Methods}
\vspace{-.2cm}

\label{sec:methods}

We first explain how we detect when recovery is needed using $M$ in Sec.~\ref{sec:ood_detection}.
We then discuss our recovery pipeline in Sec.~\ref{sec:offline_data_gen}.
Sec.~\ref{sec:project} details how we use $M$ to perform ID projection to compute a goal state for the recovery.
In Sec.~\ref{sec:offline_cmode_selection}, we discuss a method to plan contact modes that provide constraints and objective functions for the trajectory optimization.
Finally, Sec.~\ref{sec:contact_diffusion} explains how given a dataset of recovery trajectories, we can train a diffusion model to diffuse recovery optimization problems online.

\vspace{-.2cm}
\subsection{Diffusion Preliminary}
\vspace{-.1cm}
\label{app:diffusion_preliminary}
Diffusion models \cite{ddpm, song2020score} are trained to reverse a Markov process $q(x^k|x^{k-1})$ by which data $x$ is iteratively corrupted by Gaussian noise.
There are multiple ways to formulate this reverse diffusion process, one of which we discuss here to provide context for our method.
We can interpret a diffusion model as learning a function $\epsilon(x^k, k)$, where $x^k$ is a noisy input and $k$ is the step in the denoising process.
At each step of the denoising process, we update $x^k$ as follows:
\vspace{-.2cm}
\begin{equation}
    \label{eq:reverse_diff}
    x^{k-1} = x^k + \alpha_k \cdot \epsilon(x^k, k) + \mathbf{z}\sqrt{2\alpha_k}, \quad \mathbf{z} \sim \mathcal{N}(0, I)
    \vspace{-.2cm}
\end{equation}
$\alpha$ is a learning-rate hyperparameter that follows a schedule over the reverse diffusion process.
$\epsilon$ approximates the score function, or $\nabla_x \mathrm{log}~p(x)$, and the reverse diffusion is treated as sampling using Langevin Dynamics.
By performing the reverse diffusion, the learned diffusion model approximately samples from $p(x)$ by iteratively sampling from $p_\theta(x^{k-1}|x^k)$, with learned parameters $\theta$.

% Optionally, we can additionally learn functions $\epsilon(x^k, c, k)$, where $c$ is a context input.

% \vspace{-.4cm}
\subsection{OOD Detection}
% \vspace{-.3cm}
\label{sec:ood_detection}
We use the task diffusion model $M$ to detect when recovery behavior is required.
Intuitively, we seek to detect states from which $M$ is less likely to sample useful trajectories.
As diffusion models approximate probability distributions over their training data, we can estimate likelihoods of trajectories under the model to use as an out-of-distribution, or OOD, metric.
Given a state $\mathbf{s}$, we sample several trajectories from $M$ conditioned on $\mathbf{s}$ and take the average of each trajectory's likelihood.
With this, we compute a value $\tilde{p}(\mathbf{s}) \propto p(\traj | \mathbf{s})$ that estimates $M$'s ability to sample useful trajectories beginning at $\state$.
We use the trajectory likelihood calculation method from \cite{zhou2023adaptiveonlinereplanningdiffusion}, which quantifies the ability of the model to denoise a trajectory after it has been corrupted by various levels of Gaussian noise.
It does this by computing $L(\traj)=\frac{1}{|I|}\sum_{k \in K}KL(q(\traj^{k-1}|\traj^k, \traj)||p_\theta(\traj^{k-1}|\traj^k))$, where $K$ is a subset of diffusion timesteps and $KL$ is the Kullback-Leibler divergence.
The choice of $K$ is tunable, and we use $\{5, 10, 15\}$ as used in \cite{zhou2023adaptiveonlinereplanningdiffusion}.

\noindent More formally, we compute:
\vspace{-.25cm}
\begin{equation}
    \label{eq:state_likelihood}
    \tilde{p}(\mathbf{s}) = \mathbb{E}_{\traj \sim p(\traj|\mathbf{s})} L(\traj)
\vspace{-.2cm}
\end{equation}
where sampling from $M$ approximates sampling from $p(\traj|\mathbf{s})$ and $L(\traj)$ is the likelihood.
We detect OOD if $\tilde{p}(\mathbf{s})$ is less than a threshold $p_{\mathrm{min}}$.

% To provide a more interpretable OOD score that is normalized between 0 and 1, we construct an empirical cumulative distribution function (ECDF) \cite{li2022ecod} of likelihoods using $D$.
% As shown in Alg. \ref{alg:ecdf_con}, for each trajectory in $D$, we evaluate \eqref{eq:state_likelihood} for the initial state and construct a distribution over training data likelihoods.
% By querying this ECDF with the likelihood corresponding to a novel state, we can estimate if it is in-distribution.
% A lower ECDF value corresponds to less in-distribution states.

% By setting a threshold $p_{\mathrm{min}}$ of the ECDF output, we can detect when recovery is required using our likelihood estimates.

% \begin{algorithm}[h]
%     \caption{ECDF Construction}
%     \label{alg:ecdf_con}
% \begin{algorithmic}[1]
%     \Require Task dataset $D$
%     \State $L=\varnothing$
%     \For{each $\mathbf{s}_0 \in D$}
%         \State Compute $p_{\mathbf{s}_0}$ using \eqref{eq:state_likelihood}
%         \State $L = L \cup p_{\mathbf{s}_0}$
%     \EndFor
%     \State Fit ECDF to $L$
% \end{algorithmic}
% \end{algorithm}
\vspace{-.2cm}
\subsection{Offline Recovery Trajectory Generation}
% \vspace{-.3cm}
\label{sec:offline_data_gen}
In this section, we explain how we generate a dataset $D_R$ of recovery trajectories and contact modes, where $D_R = \{  (\traj_i, \mathbf{c}_i) \}_{i=1}^N$.
% $\traj$ is a trajectory and $\mathbf{c}$ is a contact mode.
% To generate a dataset of recovery trajectories, we execute the task in simulation with perturbations that can induce recovery states.
% We use Eq. \eqref{eq:state_likelihood} while executing to estimate state likelihoods and trigger recovery when the current state goes OOD.
% As shown in Fig.~\ref{fig:block_diagram}, given a recovery state $\state_R$, we then compute a trajectory to recover to a state that is ID for $M$.
% We construct a trajectory optimization problem to compute the recovery behavior, choosing the contact mode that leads to the highest likelihood state.

\begin{figure}[t]   
% \centerline{\includegraphics[scale=.65]{figures/projection.pdf}}
\centerline{\includegraphics[width=\linewidth]{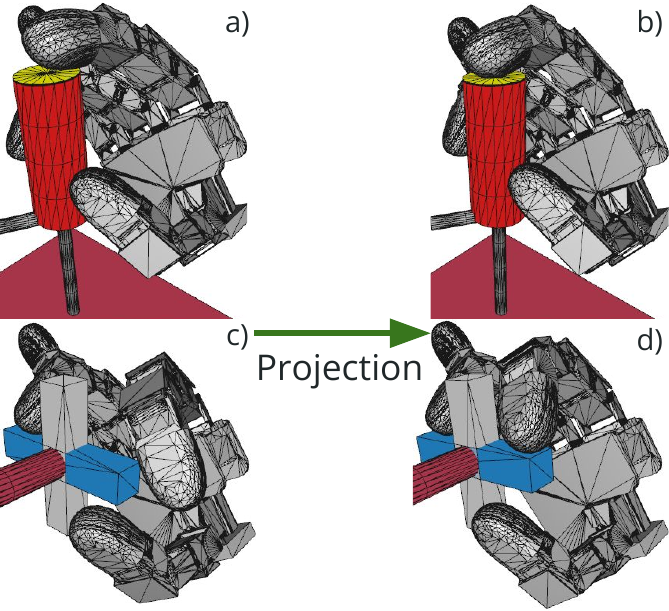}}
\vspace{-.5cm}
\caption{\textbf{a)} an OOD screwdriver turning state. \textbf{b)} a projected ID state sampled from $M$.
\textbf{c)} an OOD valve turning state. \textbf{d)} a projected ID state sampled from $M$.}
\label{fig:projection}
\end{figure}
% \vspace{-.2cm}
\subsubsection{In-Distribution State Projection}
% \vspace{-.2cm}
\label{sec:project}
We use $M$ to project an OOD state $\state_R$ ID. 
Since $M$ is trained on task executions, projecting $\state_R$ ID should result in a state from which task success is most likely, more so than possible alternatives such as attempting to recover to the most recently encountered state before recovery was initiated.
As diffusion models can be interpreted as learning a score function, we can follow the learned likelihood gradient to compute an ID goal state $\state_g$.
This is akin to sampling $\state_g$ from $M$ by indexing the initial state of a trajectory sampled from $M$.
$M$ can be conditioned on $\state_R$, sampling from $p(\traj|\state_R)$, but it can also be used to generate unconditioned trajectories, sampling from $p(\traj)$.
If there is no state conditioning, we would expect $M$ to sample high-likelihood trajectories, the initial states of which would be good ID targets.
However, directly sampling from $p(\traj)$ could produce arbitrary ID states that may be difficult to reach from $\state_R$.
We therefore modify the reverse diffusion process to sample an ID state informed by $\state_R$.

To do this, we begin the reverse diffusion process in \eqref{eq:reverse_diff} by sampling from $p(\traj|\state_R)$, but remove the $\state_R$ conditioning partway through the reverse diffusion.
% By controlling when in the process we remove the conditioning, we control how strongly $M$ projects $\state_R$.
We use the conditioned predictions for the first $T_{\mathbf{s}}$ diffusion steps, then remove the conditioning for the remaining steps.
% As $\sinit$ conditioning is commonly achieved using in-painting, as in \cite{trajdiffuser}
When $T_{\mathbf{s}}=0$, the diffusion samples from the initial state distribution in $M$'s training dataset with no influence from $\state_R$.
When $T_{\mathbf{s}}=T_D$, we obtain $\state_R$.
To compute $\state_g$, we sample $N_g$ trajectories from $M$ using the modified reverse diffusion and choose the initial state of the highest likelihood trajectory.
We show example ID projections in Fig.~\ref{fig:projection}, where the projection results in a state where the screwdriver is upright and the fingers are in more advantageous positions for turning.
For the valve, the index finger configuration is more firmly placed on the valve.

% \begin{wrapfigure}[14]{r}{0.6\textwidth} % Position 'r' for right, width of half the text width
% \vspace{-20pt} % Optional, adjust space above the figure
% \begin{minipage}{0.95\linewidth} % To prevent overflowing the wrapfigure
% \vspace{-.5cm}

% \end{minipage}
% \end{wrapfigure}

% \vspace{-.4cm}
\subsubsection{Contact Mode Planning}
% \vspace{-.2cm}

\label{sec:offline_cmode_selection}
We next need to choose a contact mode $\mathbf{c}_R$ that will determine the constraints and objective functions for the recovery trajectory optimization.
Appendix \ref{app:traj_opt} includes detail on how different contact modes are used to construct different trajectory optimization problems.
We select $\cmode_R$ using a greedy search over a set of recovery modes $C_R$ and choose the mode that results in the maximum likelihood state.
As show in Alg.~\ref{alg:offline_contact_mode_selection}, we run trajectory optimization for each mode.
% The trajectory optimizer we use in this work, CSVTO \cite{constrained_stein}, is an iterative gradient-based optimizer.
% Therefore, we can control the budget by choosing the number of optimization iterations.
We evaluate the terminal state of the optimized plan and select the contact mode that has the highest likelihood of reaching that state.
While this computation is expensive as we run the trajectory optimization for multiple modes, we can afford to spend more computation time offline when generating data.
% In Section \ref{sec:contact_diffusion}, we detail a faster method used online.

Once the contact mode is chosen, we execute the optimized trajectory.
To account for error and disturbances, we run additional optimization steps after each action is executed.
We do not perform receding horizon control, meaning the optimized trajectory reduces in length over the course of execution.
To add trajectories to $D_R$, we concatenate the executed timesteps with the remaining plan at each timestep.
This allows us to augment our data by adding a total of $H$, the horizon, trajectories to $D_R$ per recovery mode.
We only add these trajectories to $D_R$ if, after complete execution of the recovery trajectory, the likelihood of the state has increased.
This condition may not be met due to environmental perturbations.
After the recovery trajectory is fully executed, we check if the state is now ID.
If the state is not ID, we repeat the recovery process until the state is ID or until a maximum episode length is reached.
Once we have reached an ID state, we continue with task execution, executing recovery behavior when necessary.

\vspace{-.2cm}
\subsection{Trajectory Optimization Diffusion}
\vspace{-.1cm}

\begin{algorithm}[t]
    \caption{Offline Recovery Trajectory Generation}
    \label{alg:offline_contact_mode_selection}
\begin{algorithmic}[1]
    \Require OOD state $\state_R$, Recovery contact mode set $C_R$, Task diffusion model $M$, Recovery dataset $D_R$, OOD threshold $p_\mathrm{min}$, Trajectory optimizer \texttt{TrajOpt}

    \While {$\tilde{p}(\state_R) < p_\mathrm{min}$}
        \State Sample $\state_g$ from $M$, informed by $\state_R$
        
        \State $L_\cmode=\varnothing$
        \For{each $\mathbf{c} \in C_R$}

            \State $\traj = \texttt{TrajOpt}(\state_R, \state_g, \cmode)$
            \State $L_\cmode = L_\cmode \cup \tilde{p}(\mathbf{s}_H)$ ~~~~\texttt{//Final state}
        \EndFor
        \State $\mathbf{c}_R = \underset{\cmode}{\mathrm{argmax}}~L_\cmode$
        \State $\state_R' = $ state after executing $\traj$
        \State Add $(\traj, \cmode_R)$ to $D_R$ if $p(\state_R') > p(\state_R)$
        \State $\state_R = \state_R'$
    \EndWhile
\end{algorithmic}
\end{algorithm}

% \vspace{-.3cm}
\label{sec:contact_diffusion}
The method proposed in Sec.~\ref{sec:offline_data_gen} can be expensive as it requires solving multiple trajectory optimization problems to select $\cmode_R$.
To address this, we train a diffusion model that distills the recovery planning process by diffusing a recovery trajectory optimization problem.
We train a second diffusion model $M_R$ on $D_R$ that models $p(\mathbf{c}, \bm \tau|\mathbf{s}_0)$, or the joint probability of trajectories and contact modes conditioned on an initial state.
Learning this joint distribution allows us to generate the full parameterization of the trajectory optimization problem we solve to realize recovery behavior, including constraints, goal state, and initializations.
We train $M_R$ to approximate the following expression: $p(\cmode, \traj | \sinit) =p(\traj | \sinit)p(\cmode | \traj, \sinit)$, derived using the chain rule of probability.
We can interpret $M_R$ as sampling a trajectory conditioned on the initial state while simultaneously estimating the contact mode that would induce the sampled trajectory.

\begin{figure}[t]   
% \centerline{\includegraphics[scale=.7]{figures/model_diagram.pdf}}
\centerline{\includegraphics[width=\linewidth]{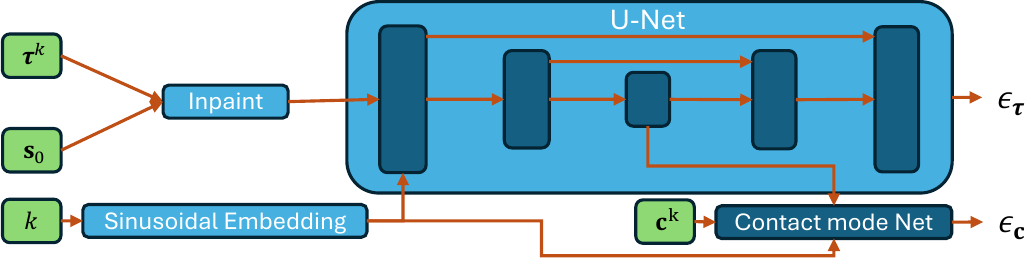}}
% \centerline{\includegraphics[scale=.95]{figures/projection_screwdriver_only.pdf}}
\vspace{-.45cm}
\caption{The architecture of the trajectory optimization diffusion model.}
\vspace{-.1cm}
\label{fig:model_diagram}
\end{figure}

Our model architecture is shown in Fig.\ref{fig:model_diagram}.
We use the U-Net Diffuser architecture \cite{trajdiffuser} to predict $\epsilon_{\traj}$, the denoising update for $\traj$.
We use a multi-layer-perceptron (MLP) that takes in the diffusion step $k$, noisy $\cmode^k$, and $\traj^k$ encoded features from the U-Net to predict $\epsilon_\cmode$, the denoising update for $\cmode$.
We use a Sinusoidal Positional Embedding \cite{vaswani2017attention} of $k$.

% \subsection{Conditioning}
% \label{app:diffusion_conditioning}
We condition the diffusion output on $\traj, \sinit$ using multiple methods.
We sample from $p(\traj | \sinit)$ using in-painting, as done in \cite{trajdiffuser}.
In-painting is a method that directly overwrites part of the diffusion sample (trajectory) with the conditioning variable (initial state).
We sample from $p(\cmode | \traj, \sinit)$ using classifier-free guidance (CFG) \cite{ho2022classifierfree}. 
which learns an additional score function $\tilde{\epsilon}(\cmode^k, k, \traj, \sinit)$ and combines the predictions of $\tilde{\epsilon}$ with $\epsilon(\cmode^k, k)$ to compute $\epsilon_\cmode$.
CFG learns $\tilde{\epsilon}$ and $\epsilon$ at training time and combines both in reverse diffusion:
\vspace{-.3cm}
\begin{equation}
    \epsilon_\cmode(\cmode^k, k, \traj, \sinit) = (1+w)\cdot\tilde{\epsilon}(\cmode^k, k, \traj, \sinit) - w\cdot\epsilon(\cmode^k, k)
    \vspace{-.2cm}
\end{equation}
$w$ is a weight that controls the strength of the conditioning.
To learn both $\epsilon_c$ and $\epsilon$, the conditioning input is randomly dropped out at training time with probability $p_\mathrm{drop}$.
As $\sinit$ is the first state of $\traj$, we combine the two variables when implementing the model into a single conditioning on $\traj$.

When recovering, we sample $N_R$ pairs of $(\traj, \cmode)$ from $M_R$, conditioned on the OOD $\state_R$.
We recover using the mode $\cmode_R$ with the highest likelihood sum.
We use the final state of the highest likelihood trajectory with contact mode $\cmode_R$ as $\state_g$.
As shown in Fig.~\ref{fig:block_diagram}, this effectively generates a trajectory optimization problem using $M_R$.
While we still run trajectory optimization online to improve constraint satisfaction, we avoid expensive contact mode selection and obtain initializations that allow us to use a smaller optimization budget.
%===============================================================================

\vspace{-.1cm}
\section{Experimental Results}
% \vspace{-.1cm}
\label{sec:results}

We evaluate on a valve turning task and a screwdriver turning task in simulation and on hardware.
We use an Allegro hand with position-based PD control.
We use CSVTO \cite{power2024constrained} for trajectory optimization.
We use CSVTO to generate task model training data, as in \cite{kumar2024diffusion}, and use the same diffusion-initialized trajectory optimization for task execution.
In all tasks, the pose of the hand is fixed.
All simulations are implemented in Isaac Gym \cite{makoviychuk2021isaac}.
All methods are evaluated for 50 trials unless otherwise specified.
Each trial corresponds to a full episode of task execution, which can include multiple recoveries and has a maximum length of 100 timesteps.

% Hyperparameter values are in Appendix~\ref{app:hyperparam}.

\vspace{-.2cm}
\subsection{Ablations and Baselines}
\vspace{-.1cm}
\label{sec:baselines}
We compare \name against the following baselines and ablations to evaluate the quality of our recovery behavior and the effect recovery has on overall task performance.
We baseline against Recovery RL \cite{thananjeyan2021recovery}, an existing safe reinforcement learning method.
This method learns a function $Q_{risk}(\mathbf{s}, \mathbf{u})$ which predicts the probability that executing action $\mathbf{u}$ at state $\mathbf{s}$ will lead to violating a pre-defined constraint.
Recovery is initiated when the prediction exceeds a threshold $\epsilon_{risk}$, and is executed using a model predictive control (MPC) method with $Q_{risk}$ as the cost, as done in \cite{thananjeyan2021recovery}.
We use MPPI \cite{mppi}, a sampling-based MPC method, as our controller for this baseline and use Isaac Gym as the dynamics model.
We train $Q_{risk}$ offline on a dataset generated by executing the task, with perturbation, as in \cite{thananjeyan2021recovery}, but with no recovery.
We compare against the method, No Recovery, which generates this dataset to motivate the use of recovery behavior.
We do not update $Q_{risk}$ online to evaluate its ability to generalize in comparison to the diffusion likelihood.

For the first ablation, Likelihood MPPI, we detect recovery scenarios using \eqref{eq:state_likelihood} and execute recovery behaviors using MPPI with the likelihood as the cost.
With this, we evaluate the efficacy of our trajectory optimization approach for generating recovery behavior, including our ID projection method.
For the second ablation, Contact Mode MLP, we evaluate the utility of diffusing $\cmode_R$ with $M_R$. 
We train an MLP to predict $\cmode_R \in \mathbb{R} ^ {n_f}$ given $\sinit$, posed as multi-label binary classification.
We sample recovery trajectories from a diffusion model conditioned on $\sinit$ and $\cmode_R$, as in \cite{kumar2024diffusion}.
This is a different diffusion model from $M_R$ but is also trained on $D_R$ as is the MLP.

We also evaluate an RL method and a contact-implicit MPC (CI-MPC) method designed to remove the need for explicit contact reasoning or reasoning about task execution vs. recovery modes.
We evaluate HORA \cite{qi2023hand}, which learns a model-free policy for in-hand object re-orientation, and IDTO \cite{kurtz2023inverse}, which performs CI-MPC on multiple platforms, including an Allegro hand.
Importantly, to align closer to the original implementations and take advantage of their ability to quickly react to disturbances, we run these methods at a higher frequency than the other methods.
We run these methods at 20x the frequency in simulation, meaning we step the simulation for 1/20 the time before re-planning.

\vspace{-.2cm}
\subsection{Metrics}
% \vspace{-.1cm}
To evaluate recovery performance, we evaluate three metrics.
The first, Recovery Success, measures how often the recovery returns the OOD $\state_R$ to an ID state, combining consecutive recovery trajectories.
The second, Recovery Drop, measures how often the robot drops the object during recovery.
The third, Recovery Timeout, measures how often the episode ends while recovery behavior is being executed.
We also evaluate on task performance metrics, which are specific to each task.
In addition, we compare the recovery planning time taken by our offline data generation and $M_R$.

\vspace{-.1cm}
\subsection{Simulated Valve Turning}
% \vspace{-.1cm}

% \begin{table}[tbp]
% \begin{center}
% \begin{tabular}{cccc}
% \Xhline{2pt}
% \multirow{3}{*}{Method}& \multirow{3}{*}{\parbox{1.5cm}{\centering Recovery Success }$\uparrow$}  & \multirow{3}{*}{\parbox{1.5cm}{\centering Recovery Timeout }$\downarrow$}&\multirow{3}{*}{\parbox{1.5cm}{\centering Distance to goal (rad.) }$\downarrow$}\\
% & & & \\
% & & & \\
% \hline
% \name (Ours)& 93.0\%& 7.0\%& $\mathbf{.20 \pm .06}$\\
% \namenospace-MLP Ablation& 85.0\%&  15.0\%& $.38 \pm .15$\\
% Recovery RL &  \textbf{94.0\%}& \textbf{6.0\%}& $1.57 \pm .20$\\
% Likelihood MPPI&  0\% & 100\% & $.75 \pm .15$\\
% No Recovery & - & -  &$.67 \pm .14$\\
% \hline
% \end{tabular}
% \end{center}
% \caption{Valve Performance Statistics with 95\% confidence interval for distance to goal.
% There is no possible dropping for the valve task.}
% \label{tab:valverecoveryresults}
% \end{table}

\begin{figure}[t]
% \vspace{-1.4cm}
\centering
\includegraphics[width=\linewidth]{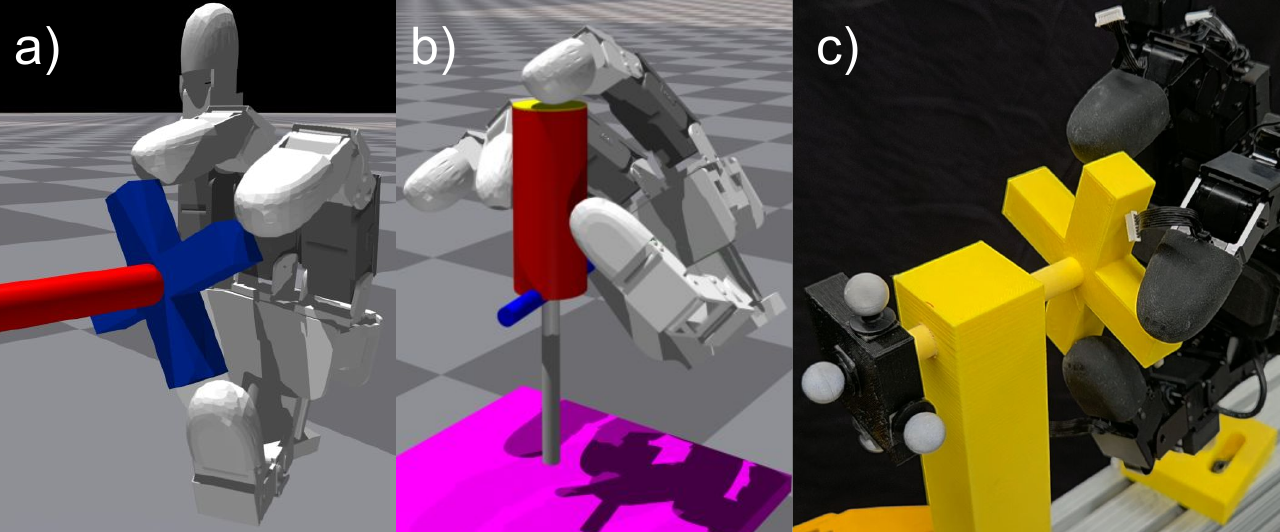}
\vspace{-.7cm}
% \includegraphics[width=\linewidth]{figures/task_environments.pdf}
% \vspace{-.6cm}
\caption{\textbf{a)} Simulated valve. \textbf{b)} Simulated screwdriver. \textbf{c)} Hardware valve.}
% \vspace{-.1cm}
% \includegraphics[scale=.6]{figures/screwdriver_environment_vert.pdf}
% \caption{Screwdriver turning task}
\label{fig:tasks}
\end{figure}

In this task, shown in Fig.~\ref{fig:tasks}a, the robot turns a simulated valve clockwise using the thumb, index, and middle fingers.
The low friction small surface area of the valve necessitate precise finger placement to execute turning.
The position of the valve is fixed and it rotates in 1 dimension.
$\obj$ is the valve's orientation.
To execute the task, we sequence primitives which each attempt to turn the valve $-\frac{\pi}{4}$ radians while perturbing executed actions.
A $-\frac{\pi}{4}$ rotation may not be achievable due to perturbations and execution error.
The overall task goal is to turn the valve by $-\frac{\pi}{3}$ radians from the random initial configuration.
% The initial valve angle is uniformly initialized for each trial in the range $[-\frac{\pi}{4}, \frac{\pi}{4})$.
% The initial robot configuration is also randomly initialized by adding Gaussian noise with standard deviation .06 to each joint angle of a nominal configuration.
During execution, we perturb the actions with Gaussian noise with standard deviation .03, further increasing the challenge of the fine manipulation task.
% We collect a dataset of executions to train $M$.
% We filter the dataset to only include turning trajectories which turn the valve at least $-\frac{\pi}{8}$ radians.

We use three recovery modes, where each mode resets 1 finger while the other 2 remain in contact with the valve.
% Using these recovery modes, we can reset fingers that are not in a position on the valve conducive to turning.
For this task, the task performance metric measures distance to the $\frac{\pi}{3}$ rad. goal.
We run the No Recovery baseline for 179 trials.
The Recovery RL constraint is violated if any finger breaks contact with the valve.

\vspace{-.2cm}
\subsection{Simulated Screwdriver Turning}
\vspace{-.1cm}
In this task, shown in Fig.~\ref{fig:tasks}b,  the robot turns a simulated precision screwdriver clockwise using the thumb, index, and middle fingers.
The base of the screwdriver is fixed to the table but is free to rotate in 3 dimensions, simulating driving a screw in a slot.
$\obj$ is the 3D orientation of the screwdriver, parameterized by roll, pitch, and yaw Euler angles.
The blue stalk on the screwdriver is for visualization only, meaning the screwdriver is symmetric about its yaw-axis.
The goal of the task is to turn the screwdriver clockwise as far as possible.
To do this, we sequence primitives that each attempt to turn the screwdriver by $-\frac{\pi}{2}$ radians, though this may not be achieved due to environmental perturbations.
% We collect a dataset of executions to train $M$.
% We filter the dataset to only include turning trajectories which turn the valve at least $-\frac{\pi}{6}$ radians.
We randomly initialize the object configuration and perturb the system during task execution by applying random wrenches to the screwdriver.

% We sample initial roll and pitch angles from $\mathcal{N}(0, .01)$ and yaw angles uniformly in the range $[-\frac{\pi}{4}, \frac{\pi}{4})$. 
In simulation, we apply a random wrench at each action step with probability $\frac{1}{3}$.
When a wrench is applied, we first randomly sample a point on the screwdriver body and apply a force perpendicular to the surface of the screwdriver, with a random rotation about the yaw axis of the screwdriver in the range of $[-\frac{\pi}{10}, \frac{\pi}{10}]$.
Force magnitudes are uniformly randomly sampled from the range $(0, 1.5)$.

We use two recovery modes: (1) in which the thumb and middle fingers are reset, and (2) in which the index finger is reset.
Non-resetting fingers remain in contact during recovery.
These recovery modes are inspired by human use of precision screwdrivers.
We account for the screwdriver's yaw-axis symmetry by enforcing that $\state_g$ has the same yaw as $\state_R$ before initiating recovery.
This focuses recovery on the non-symmetric components of the state that have an effect on future task performance.
For this task, we run the No Recovery baseline for 130 trials to generate data for the Recovery RL baseline.
The constraint for Recovery RL is violated if the screwdriver is dropped.

\vspace{-.2cm}
\subsection{Hardware Experiments}
\vspace{-.1cm}

\begin{table}[tbp]\footnotesize
\begin{center}
\begin{tabular}{cccccc}
\Xhline{2pt}
&\multirow{3}{*}{\parbox{1cm}{\centering Method}}& \multirow{3}{*}{\parbox{1.1cm}{\centering Recovery Success }$\uparrow$} & \multirow{3}{*}{\parbox{1.1cm}{\centering Recovery Drop }$\downarrow$} & \multirow{3}{*}{\parbox{1.1cm}{\centering Recovery Timeout }$\downarrow$}&\multirow{3}{*}{\parbox{1cm}{\centering Task Metric (rad.) }$\downarrow$}\\
& & & & & \\
& & & & & \\
\hline
\multirow{7}{*}{Sim.Valve}& \name (Ours)& 93.0\%& - & 7.0\%& $\mathbf{.20 \pm .06}$\\
& \namenospace-MLP& 85.0\%& - & 15.0\%& $.38 \pm .15$\\
& Recovery RL \cite{thananjeyan2021recovery} &  \textbf{94.0\%}& - &\textbf{6.0\%}& $1.57 \pm .20$\\
& Likelihood MPPI&  0\% & - &100\% & $.75 \pm .15$\\
& No Recovery & - & -  & -&$.67 \pm .14$\\
& HORA \cite{qi2023hand}&  - & - &-  & $ .32 \pm .08$\\
& IDTO \cite{kurtz2023inverse} & - & -  & -& $ .51\pm .19$\\
\hline
\multirow{7}{*}{\parbox{1.5cm}{\centering Sim. Screwdriver }}&\name (Ours)& \textbf{78.9\%}& 16.8\%& 4.3\%& $\mathbf{-1.51 \pm .27}$\\
&\namenospace-MLP& 78.5\% & \textbf{16.3\%} & 5.1\%& $-1.42 \pm .24$\\
&Recovery RL \cite{thananjeyan2021recovery} & 69.7\% & 27.3\%& 3.0\%& $-1.19 \pm .09$\\
&Likelihood MPPI& 3.9\%  &96.1\% & \textbf{0.0\%} &$-.46\pm .05$\\
&No Recovery & - & - & - & $-1.15 \pm .08$\\
& HORA \cite{qi2023hand}&  - & - &-  & $ -.46 \pm .24$\\
& IDTO \cite{kurtz2023inverse} & - & -  & -&$ -.62\pm .08$\\
\hline
\multirow{3}{*}{\parbox{1.5cm}{\centering Hardware Valve }}&\name (Ours)& 89.3\%& -& 10.7\%& 
$\mathbf{.18 \pm .10}$\\
&Recovery RL \cite{thananjeyan2021recovery} & \textbf{93.2\%} & - & \textbf{6.8\%} & $ 1.3 \pm .17$\\
& Likelihood MPPI & 0\% & - & 100\% & $.77 \pm .20$\\
\hline
\multirow{3}{*}{\parbox{1.5cm}{\centering Hardware Screwdriver }}&\name (Ours)& \textbf{63.2\%}& \textbf{36.8\%}& \textbf{0\%}& 
$\mathbf{-1.08 \pm .58}$\\
&Recovery RL \cite{thananjeyan2021recovery} & - & - & - & $-.55 \pm .35$\\
& Likelihood MPPI & 0\% & 100\% & \textbf{0\%} & $-.58 \pm .1$\\
\hline

\end{tabular}
\end{center}
\vspace{-.3cm}
\caption{95\% confidence interval for task metric. Dropping is not possible for the valve task. Task metric: distance to goal for valve task, rad. turned for screwdriver.}
\label{tab:recoveryresults}
\vspace{-.4cm}
\end{table}

We run 10 trials of \namenospace, Recovery RL, and Likelihood MPPI on hardware versions of the screwdriver task shown in Fig.~\ref{fig:TODR_title_figure}.
We use a Vicon motion capture system to estimate $\obj$.
To avoid issues with hardware overheating for the screwdriver task as a result of the higher forces needed for manipulation, we run each episode for a maximum of 50 steps.
As in simulation, we randomly sample wrench perturbations with probability $\frac{1}{3}$ each timestep.
If a perturbation is applied, the location of the perturbation is randomly chosen as either pushing the screwdriver toward the palm or perpendicular to the palm, with the applied force pointing toward the thumb.
The perturbation is executed by a human (paper author) poking the screwdriver with a wooden rod.
For the valve task, we run 20 trials of the same methods and do not add Gaussian perturbation to the actions.

% \begin{table}[tbp]
% \begin{center}
% {\begin{tabular}{ccccc}
% \Xhline{2pt}
% \multirow{3}{*}{Method}& \multirow{3}{*}{\parbox{1.5cm}{\centering Recovery Success }$\uparrow$} & \multirow{3}{*}{\parbox{1.5cm}{\centering Recovery Drop }$\downarrow$} & \multirow{3}{*}{\parbox{1.5cm}{\centering Recovery Timeout }$\downarrow$}&\multirow{3}{*}{\parbox{1.5cm}{\centering Turn Amount (rad.) }$\downarrow$}\\
% & & & & \\
% & & & & \\
% \hline

% \hline
% \end{tabular}}
% \end{center}
% \caption{Hardware Screwdriver Performance Statistics with 95\% confidence interval for turn amount}
% \label{tab:hardware_results}
% \end{table}

\vspace{-.2cm}
\subsection{Results}
\vspace{-.1cm}
As shown in Table~\ref{tab:recoveryresults}, \name leads to better task performance than the baselines, with comparable to better performance to the ablation.
Our results suggest recovery improves task performance as our method outperforms the No Recovery baseline on both tasks.
We find that even though the MLP Ablation is capable of recovering, it can sometimes predict contact modes not seen in $D_R$.
We find that the ablation predict contact modes for the valve without any resetting fingers even though all contact modes in $D_R$ have a finger being reset.
\name only selects contact modes that appear in $D_R$ and the task performance metrics suggest it learns a more useful recovery trajectory distribution.

The Recovery RL baseline has a high recovery success rate for the valve task, indicating that $Q_{risk}$ can be used to guide MPPI.
However, this high recovery rate does not translate to task performance. This is possibly due to the poor generalization of $Q_{risk}$ or its failure to encode knowledge of task performance, unlike the likelihood.
The sensitivity of $Q_{risk}$ combined with the difficulty of sampling contact-constraint satisfying trajectories in MPPI leads it to drop the screwdriver more than \namenospace.

We believe the likelihood, while useful to detect recovery, is less useful as a cost.
This is reasonable as diffusion models learn highly non-linear functions to maximize the likelihood.
Sampling based MPC function can struggle to optimize given this cost landscape, leading to a high drop rate for the screwdriver task.
This motivates the use of our trajectory optimization with ID projected goal.

We find that HORA learns a useful policy for the valve task, but struggles with the screwdriver task.
This could be due to the increased complexity of the screwdriver task, as maintaining stable contact while experiencing external disturbances is important to prevent catastrophic task failure.
As RL methods rely on random sampling for exploration, this can lead to breaking contact and dropping the object.

While IDTO is capable of completing the valve task, it can fall into local minima that cause task failure. 
This leads to the higher distance to goal and larger variance in the goal distance.
While \name also uses non-convex trajectory optimization, we impose stricter constraints than IDTO that come from our diffused contact modes.
This leads to a smaller search for the trajectory optimization, which can lead to overall better solutions.
For the screwdriver task, we find that IDTO's performance, both quantitatively on task performance and qualitatively on how likely optimized trajectories are to contain contact switches, is sensitive to the tuning of parameters, and the parameter tuning that yielded the best overall task performance was observed to perform very little contact switching.
This suggests that choosing contact switches when executing a sensitive task under perturbation is a challenging problem, which benefits from explicitly planning contact modes.

We also find that \name plans resets of the thumb and middle fingers without which the robot may eventually reach kinematic limits and be unable to turn further, allowing us to not only recover from perturbations but also naturally extend task execution for repetitive tasks.
Along with recovering from execution error in the valve task, we find \name can improve execution in situations where the initial grasp of the object is not conducive to turning, as in Fig.~\ref{fig:tasks}a.
These states have low likelihoods under the task model, triggering recovery behavior to states from which turning is easier.

\name turns the screwdriver further than the baselines for the hardware task.
Recovery RL fails to detect that recovery is necessary, possibly due to poor generalization of $Q_{risk}$ leading to poor sim-to-real transfer.
In contrast, both the task model for detection and the trajectory optimization diffusion model for recovery generalize better to hardware.
This is potentially due to the diffusion model being exposed to noised data in training, improving generalization.
Likelihood MPPI detects recovery as it uses the same detection as \namenospace, but fails to recover without dropping the screwdriver.
We observe a higher drop percentage and lower recovery rate in hardware when compared to simulation, potentially due to the sim-to-real gap.
Specifically, there could be error in the modeling of the robot and object in terms of their geometries and physical properties such as friction and joint damping.
\name achieves lower distance to goal and better recovery metrics for the hardware valve task.
We achieve comparable results between simulation and hardware for \namenospace, showing less sensitivity to potential geometry or physical property mismatch for this task.

% \begin{table}[h]
% \begin{center}
% \begin{tabular}{cccc}
% \Xhline{2pt}
% &  & \multirow{2}{*}{\parbox{1.5cm}{\centering Recovery Planning}} & \multirow{2}{*}{\parbox{1.5cm}{\centering Trajectory Update}}\\
% & & & \\
% \hline
% \multirow{2}{*}{\parbox{2.5cm}{\centering Sim. Valve}}& Data Generation& 90.4& 8.5\\
% & Diffusion&  7.0& 9.0\\
% \hline
% \multirow{2}{*}{\parbox{2.5cm}{\centering Sim. Screwdriver}}&Data Generation  &22.7 & 9.5\\
%  & Diffusion &  8.2& 11.1\\
% \hline
% \end{tabular}
% \end{center}
% \caption{Recovery Planning Times (seconds)}
% \label{tab:recovery_plan_time}
% \end{table}
% Planning times are reported for each recovery trajectory of length $H$. Recovery planning represents time between OOD detection and executing the first recovery action, while trajectory update time includes all additional optimization done to the recovery trajectory over the remaining $H-1$ timesteps. We achieve faster recovery planning using our diffusion model than when generating data.

\name also enables faster planning.
For the valve task, planning a recovery trajectory takes 90.4 s for the valve task and 22.7 s for the screwdriver task during offline data generation, with additional trajectory optimization time during task execution of 8.5 s for the valve and 9.5 s for the screwdriver.
However, planning a recovery trajectory with $M_R$ takes 7.0 s for the valve task and 8.2 s for the screwdriver task, with additional trajectory optimization time during task execution of 9.0 s for the valve and 11.1 s for the screwdriver.
\vspace{-.6cm}
\section{Conclusion}
\vspace{-.1cm}
\label{sec:conclusion}

We presented \namenospace, a method to generate contact-rich recovery behaviors informed by diffusion model likelihood estimation.
Our method uses a diffusion model trained on task data to detect when perturbations or execution error require recovery behaviors and generates a dataset of these recovery trajectories.
We train a diffusion model on this dataset to more efficiently generate and solve recovery trajectory optimization problems online.
We show our method outperforms a reinforcement learning baseline and methods that do not explicitly reason about contact interactions, including on a hardware screwdriver turning task.
In future work, we plan on combining our work on planning and trajectory optimization with feedback control methods to improve execution time as well as relaxing the assumption of a prescribed set of contact modes by inferring the set from human demonstrations or attempting to learn a set of useful modes from object interaction.

\vspace{-.2cm}
\appendix
\vspace{-.1cm}

\subsection{Trajectory Optimization}
% \vspace{-.2cm}
\label{app:traj_opt}
Our trajectory optimization formulation builds on prior work from \cite{kumar2024diffusion, fanrolling}.
We partition the trajectory into $\bm{\tau} = \{\bm{\tau}_c, \bm{\tau}_r, \bm{\tau}_o\}$, where $\traj_c$ is the trajectory for contact fingers, or fingers with contact mode 1, $\traj_r$ is the trajectory for fingers being reset, or fingers with contact mode 0, and $\traj_o$ is the trajectory for the object.
Our full trajectory optimization problem is written as:

\vspace{-.6cm}
\begin{equation}
\begin{aligned}
& \min_{\substack{\mathbf{s}_1, \mathbf{s}_2, \cdots, \mathbf{s}_H; \\ \mathbf{u}_1, \mathbf{u}_2, \cdots, \mathbf{u}_H}}  J_{g}(\bm{\tau_o}) + J_{r}(\bm{\tau_r}, \bm{\tau_o}) + J_{smooth}(\bm{\tau}) \\
\end{aligned}
\vspace{-.6cm}
\end{equation}

\begin{align}
\text{s.t.} \quad & \mathbf{q}_{min} \leq \mathbf{q}_t \leq \mathbf{q}_{max} \label{q_bounds}\\
& \mathbf{u}_{min} \leq \mathbf{u}_t \leq \mathbf{u}_{max} \label{u_bounds}\\
& f_{contact}(\mathbf{s}_{c,t}, \mathbf{s}_{o,t}) = 0 \label{contact}\\
& f_{kinematics}(\mathbf{s}_{c,t}, \mathbf{s}_{o,t}, \mathbf{s}_{c, t+1}, \mathbf{s}_{o, t+1}) = 0 \label{kinematics}\\
& f_{balance}(\mathbf{s}_{c, t}, \mathbf{s}_{o, t},\mathbf{s}_{c, t+1}, \mathbf{s}_{o, t+1}, \mathbf{u}_{c,t}, \mathbf{u}_{o,t}) = 0 \label{balance}\\
& f_{friction}(\mathbf{s}_{c,t}, \mathbf{s}_{o,t}, \mathbf{u}_{c,t}) \leq 0 \label{friction}\\
& f_{min\_f}(\mathbf{s}_{c,t}) \geq f_{min} \label{min_force}\\
& f_{contact}(\mathbf{s}_{r,t}, \mathbf{s}_{o,t}) \leq - \delta , \quad t < H \label{reset}\\
& f_{contact}(\mathbf{s}_{r,H}, \mathbf{s}_{o,H}) = 0 \label{terminal_contact}\\
& f_{contact\_patch}(\mathbf{s}_{r,H}, \mathbf{s}_{o,H}) \leq r \label{contact_patch}\\
& \mathbf{q}_{r, t} + \Delta \mathbf{q}_{r,t} - \mathbf{q}_{r, t+1} = 0 \label{freespace_dynamics}
\end{align}
The cost term $J_g$ encourages the object configuration to reach $\obj_g$, or the object state component of $\state_g$.
% When generating $D_R$, we compute $\state_g$ using our projection method and diffuse it using $M_R$ online.
% Specifically, we sample $N_R$ pairs of contact modes and trajectories from $M_R$ and select $\cmode_R$ as the contact mode with the highest summed likelihood.
$\state_g$ is the final state of the highest likelihood trajectory with diffused contact mode $\cmode_R$.
% $J_r$ is a cost on the point on each resetting finger used to make contact with the object.
% Specifically, it is a cost on the distance to target $\mathbb{R}^3$ contact points for the fingers being reset.
% The target contact points are computed using $\state_g$ and are the closest points on the robot to the object.
$J_r$ incentivizes trajectories to use the same contact point on the robot after resetting as used in $\state_g$.
$J_{smooth}$ incentivizes a smooth trajectory and is the same for all contact modes.

\eqref{q_bounds}, \eqref{u_bounds} specify bounds on the robot configurations and actions respectively and are the same for all contact modes.
\eqref{contact}, \eqref{kinematics}, \eqref{balance}, \eqref{friction}, \eqref{min_force} are the same as those in \cite{fanrolling} and are used for the contact fingers.
\eqref{contact} ensures the finger stays in contact during the trajectory, \eqref{kinematics} ensures the velocity of the contact point on the robot equals the velocity of the contact point on the object, \eqref{balance} ensures the applied forces and gravity are balanced, \eqref{friction} ensures applied forces lie within a Coulomb friction cone, and \eqref{min_force} specifies a minimum force magnitude applied by each finger which is useful for overcoming unmodeled friction or other reaction forces.

\eqref{reset}, \eqref{terminal_contact}, \eqref{contact_patch}, \eqref{freespace_dynamics} are used for the fingers being reset.
\eqref{reset}, \eqref{terminal_contact}, \eqref{freespace_dynamics} are the same as in \cite{kumar2024diffusion}.
\eqref{reset} ensures that the fingers being reset avoid contact with a distance threshold $\delta$ until the final time step.
\eqref{terminal_contact} ensures that at the final step, the fingers are in contact.
\eqref{freespace_dynamics} ensures that configurations and actions are consistent for fingers not in contact.
In this work we add an additional constraint \eqref{contact_patch}, which specifies where on the object each finger should make contact when reset.
We constrain the contact point to be within a radius $r$ of the closest point on the object for each finger in $\mathbf{s}_g$.

\bibliographystyle{IEEEtran}
\bibliography{ref.bib}  % .bib

\end{document}